\documentclass[journal]{IEEEtran}
\usepackage{amsmath,amsfonts}
\usepackage{algorithmic}
\usepackage{algorithm}
\usepackage{array}
\usepackage[caption=false,font=normalsize,labelfont=sf,textfont=sf]{subfig}
\usepackage{textcomp}
\usepackage{stfloats}
\usepackage{url}
\usepackage{verbatim}
\usepackage{graphicx}
\usepackage{threeparttable}
\usepackage{cite}
\usepackage{xcolor}
\usepackage{booktabs}
\usepackage{amssymb}
\usepackage{soul}
\soulregister\cite7
\soulregister\ref7
\begin{document}
\title{Hierarchical Deformation Planning and Neural Tracking for DLOs in Constrained Environments}

\author{$\text{Yunxi Tang}$,
        $\text{Tianqi Yang}$,
        $\text{Jing Huang}$,
        $\text{Xiangyu Chu}$,
        and $\text{Kwok Wai Samuel Au}$
\thanks{
This work was supported in part by Multi-scale Medical Robotics Center, AIR@InnoHK and in part by Direct Grants (The Chinese University of Hong Kong) under Grant 4055245.
(\textit{Corresponding author: Xiangyu Chu}). 

All authors are with the Department of Mechanical and Automation Engineering, The Chinese University of Hong Kong, Hong Kong SAR, and with the Multi-scale Medical Robotics Centre, Hong Kong SAR (e-mail: yunxitang@cuhk.edu.hk; tianqiyang@cuhk.edu.hk; huangjing@mae.cuhk.edu.hk; xiangyuchu@cuhk.edu.hk; samuelau@cuhk.edu.hk).
}
}


\maketitle

\begin{abstract}
Deformable linear objects (DLOs) manipulation presents significant challenges due to DLOs' inherent high-dimensional state space and complex deformation dynamics. The wide-populated obstacles in realistic workspaces further complicate DLO manipulation, necessitating efficient deformation planning and robust deformation tracking. In this work, we propose a novel framework for DLO manipulation in constrained environments. This framework combines hierarchical deformation planning with neural tracking, ensuring reliable performance in both global deformation synthesis and local deformation tracking.
Specifically, the deformation planner begins by generating a spatial path set that inherently satisfies the homotopic constraints associated with DLO keypoint paths. Next, a path-set-guided optimization method is applied to synthesize an optimal temporal deformation sequence for the DLO. 
In manipulation execution, a neural model predictive control approach, leveraging a data-driven deformation model, is designed to accurately track the planned DLO deformation sequence.
The effectiveness of the proposed framework is validated in extensive constrained DLO manipulation tasks.
\end{abstract}

\section{Introduction}
{Deformable linear object (DLO) manipulation is common in industrial and domestic settings. Their flexibility and underactuation make precise DLO manipulation challenging~\cite{dom_chan}. Effective DLO manipulation is critical in practical applications, such as cable packing~\cite{WanyuMa2023}, cable harness assembly~\cite{Zhang2024}, constrained wiring~\cite{Galassi2021}, and surgical robotics~\cite{Joglekar2025}, where obstacle-aware planning and accurate deformation tracking are essential. While prior works~\cite{David2016, XiangLi2023, Yang2023} focus on free-space manipulation, real-world tasks often involve dense obstacles. Recent studies~\cite{Dale2020, Jinghuang2023,Jinghuang2024,Mingrui2024} consider constrained DLO manipulation through planning and tracking, yet they either assume simplified tasks~\cite{Jinghuang2023} or require dense deformation sampling~\cite{Mingrui2024}. These challenges motivate the need for efficient strategies for constrained DLO manipulation.}
\subsection{DLO Deformation Planning}
In constrained environments, local deformation control often becomes trapped by nearby obstacles due to limited prediction 
\begin{figure}[tb]
    \centering
    \includegraphics[width=0.95\linewidth]{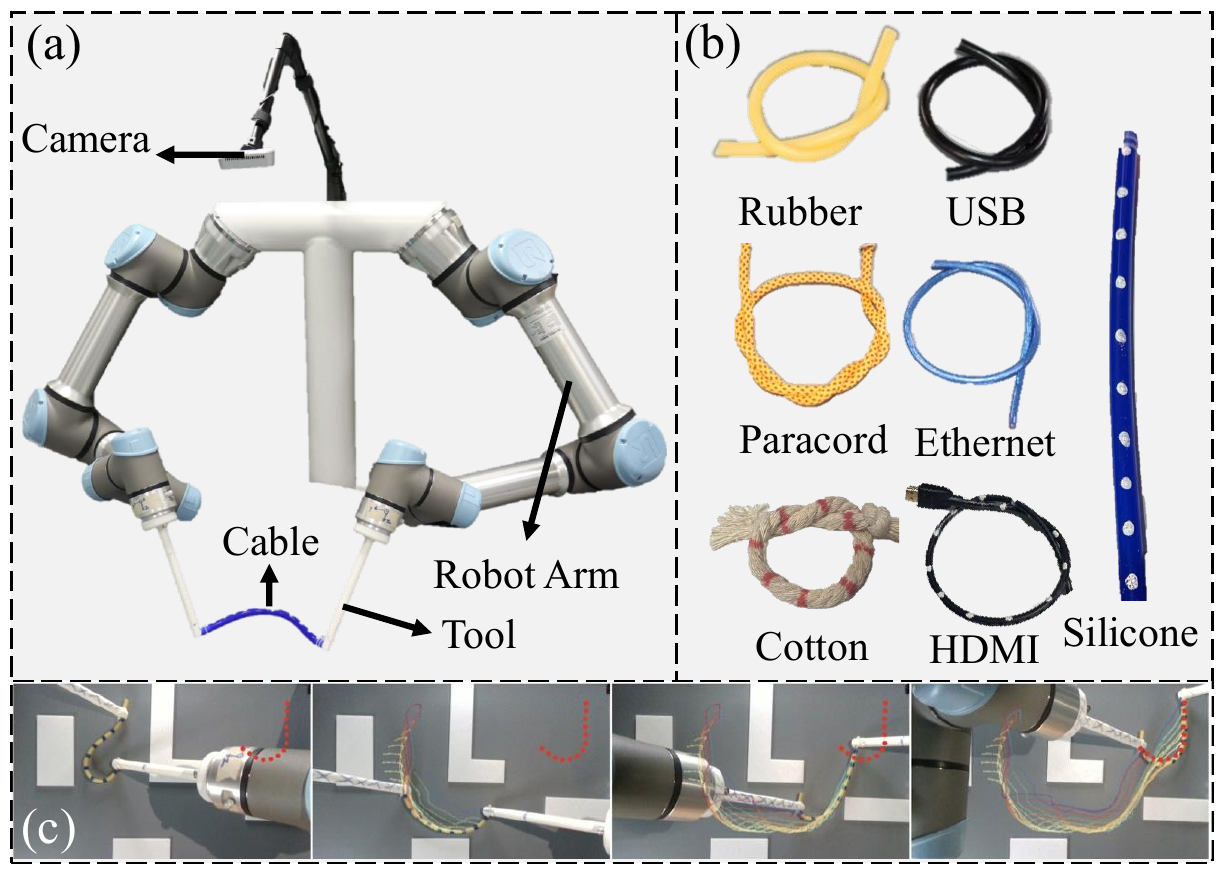}
    \caption{(a) DLO manipulation with a bi-manual robot. (b) Real-world cables. (c) Example of constrained DLO manipulation.}
    \label{fig:cables}
\end{figure}
horizon, highlighting the need for global deformation planning. {A first line of work studies model-based deformation planners~\cite{Li2020,Frank2011}, which generate feasible DLO trajectories by optimizing robot actions using forward simulations of explicit deformation models, either physics-based~\cite{Li2020} or data-driven~\cite{Tang2024}. Such approaches are computationally prohibitive in constrained environments. Each planning step requires repeated nonlinear deformation simulations, making real-time and long-horizon planning difficult. 
Some other works~\cite{Mingrui2024,Aksoy2025} utilize sampling-based planners, e.g., Rapidly-exploring Random Tree (RRT), to plan a DLO shape sequence connecting the initial and target shapes. However, raw shape samples drawn from the high-dimensional DLO state space, often consisting of more than $10$ points~\cite{Mingrui2024}, rarely correspond to physically feasible shapes. Shape post-processing such as equilibrium searches or energy minimization~\cite{Bretl2013,Mingrui2024,Moll2006} are required. For instance, a discrete elastic rod model is employed with dense sampling for 3D constrained planning in~\cite{Mingrui2024}. However, the combination of shape sampling and costly post-processing at each query makes such methods computationally inefficient and difficult to scale to complex tasks.
Another paradigm relevant to our work is spatial path set planning method for deformable objects~\cite{Jinghuang2023,Jinghuang2024} and multi-agent path finding~\cite{Swarm2018,Mao2023}. Instead of directly sampling full DLO shapes, a homotopic path set for DLO keypoints is planned in~\cite{Jinghuang2023}, which ensures the collective properties to satisfy the spatial and topological constraints of DLOs. Though spatial path set planning is more efficient than dense DLO shape sampling, such methods~\cite{Jinghuang2024} only provide coarse motion references and lack physically plausible deformation information for precise deformation regularization. Independent keypoint tracking in confined workspaces can lead to incoherent deformations such as over-compression.

In this study, to address these challenges, we introduce a hierarchical deformation planner that decouples the problem into two stages: spatial path set planning and deformation optimization. To reduce the search space of DLO shapes, we first plan a set of collision-free spatial keypoint paths that leverage the DLO’s internal kinematic constraints. Since individual paths lack physically plausible deformation details, we further propose a novel deformation generation module. This module leverages an analytical mass-spring (MS) model to refine the full DLO deformation sequence conditioned on the planned paths. Together, the hierarchical planning strategy enables efficient and physically consistent deformation planning for constrained DLO manipulation.}

\subsection{DLO Deformation Tracking}
{Reliable deformation tracking in obstacle-populated environments is essential for executing planned DLO trajectories. A common approach is Jacobian-based method~\cite{David2016,XiangLi2023,Yang2023}, which employs a linear controller to reduce shape errors using deformation Jacobians. The numerical Jacobians, reflecting the local relationship between robot motions and DLO deformations, can be estimated online~\cite{Broyden}.
However, Jacobian-based controllers often fail in cluttered environments, as the local linear approximations limit the ability to account for obstacles. To overcome the limitations of local Jacobian-based controllers, recent works adopt learning-based approaches that use deep neural network to model DLO deformations~\cite{Caporali2024,Changhao2022}. The learned deformation models can be integrated within model predictive control (MPC) frameworks to enable obstacle-aware deformation tracking. A neural deformation model is integrated into an MPC framework to achieve constrained DLO manipulation~\cite{Tang2024}. 
MPC tracking performance hinges on accurate learned deformation models. 
Recent works have explored different paradigms to improve deformation model learning for DLOs. Several studies explored DLO deformation prediction through learned latent-space dynamics~\cite{DeformNet2024,Zhang2021}. For example, a linear latent model is proposed to predict rope deformation from images~\cite{Zhang2021}. These methods require training heavy image encoder/decoders to reconstruct observations, making them computationally expensive and sensitive to variations in lighting and background conditions. Within keypoint-based approaches, graph neural networks (GNNs) incorporate physical priors by modeling local interactions among DLO keypoints~\cite{Changhao2022,Ai2025}. However, the reliance on local neighborhood aggregation struggles to capture the long-range dependencies required for large global deformations. To build long-range dependencies across the entire DLO, a bidirectional Long Short-Term Memory (Bi-LSTM) network was proposed in~\cite{lstm2025}. However, Bi-LSTM is less parallelizable and leads to slower training and inference for long sequences and real-time applications. Recent self-attention based models capture complex deformation patterns~\cite{Tang2024}, but the exclusion of EEF rotations limits its generalization in challenging tasks.

To address these limitations, we develop a neural deformation tracking controller built on learning-based MPC to track planned deformations. A novel deformation model based on an encoder-decoder Transformer architecture~\cite{attention} is proposed to approximate deformation dynamics by framing it as a sequence-to-sequence problem. By aggregating information across all vertices, the proposed deformation model jointly represents fine-grained local interactions and global shape evolution under robot actions, enabling accurate predictions over long planning horizons and large deformations. The neural tracking controller then leverages gradient-based optimization to compute robot actions that track reference DLO configurations while navigating obstacles.}

\begin{figure*}[t]
    \centering
    \includegraphics[width=0.9\textwidth]{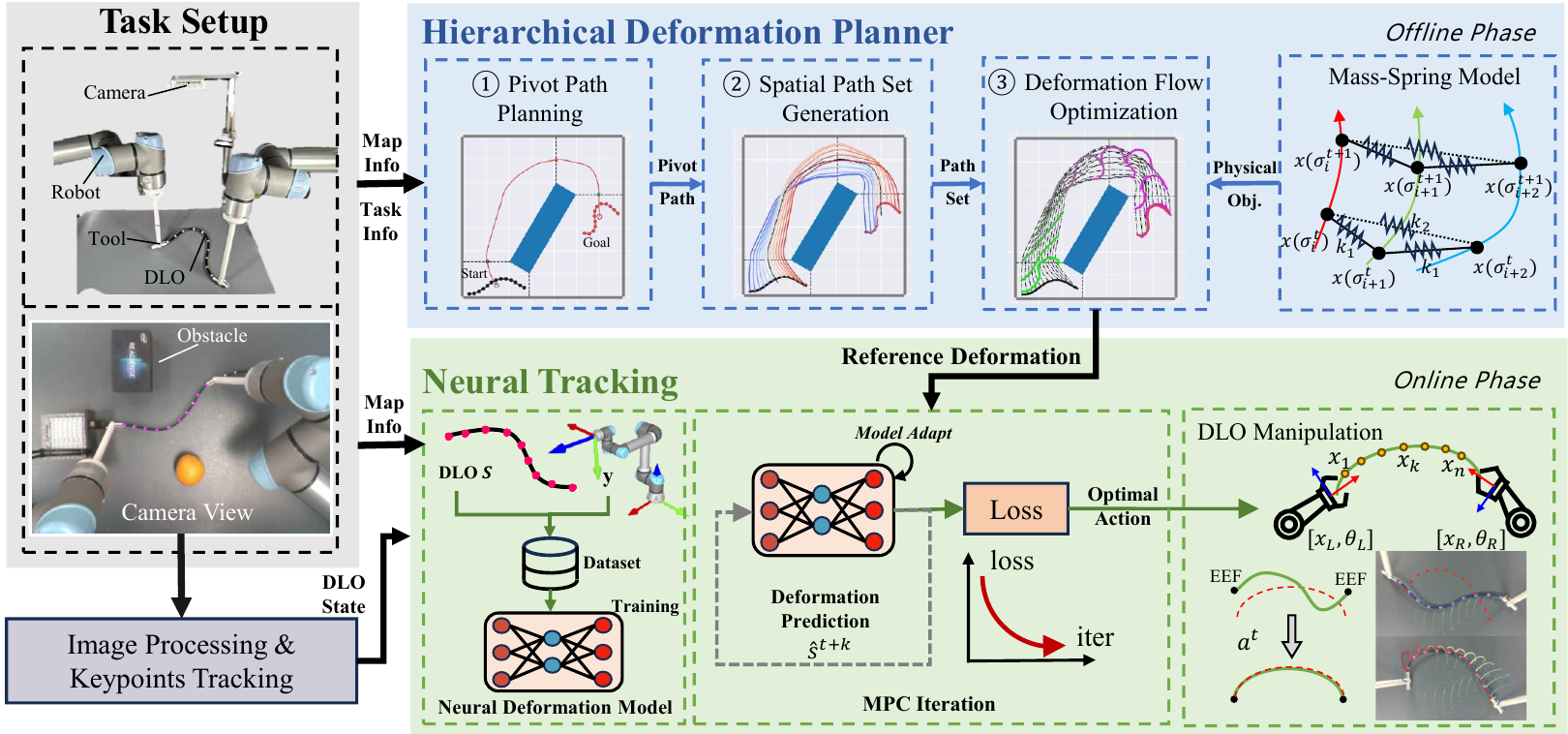}
    \caption{Pipeline of the proposed manipulation framework, unifying hierarchical deformation planning and closed-loop neural MPC tracking.}
    \label{fig:sys_pipline}
\end{figure*}

Figure~\ref{fig:sys_pipline} illustrates the proposed framework, whose key contributions are summarized as follows:

1) A hierarchical deformation planner is proposed for deformation planning, which can effectively plan global deformation sequences in obstacle-dense workspaces.

2) A neural MPC tracking controller is developed to follow the deformation flow. The controller leverages a novel Transformer-based deformation model to capture the complex interactions between robots and DLOs.

3) The proposed pipeline is validated through extensive simulation and real-world experiments, demonstrating its effectiveness in diverse constrained scenarios.

\section{Problem Statement}
This work focuses on bimanual DLO manipulation tasks on a tabletop with dense obstacles, where the DLO is grasped by end-effectors (EEFs) at both ends. Considering a DLO with $n$ feedback feature points, the state $\mathbf{s} \in \mathcal{S}$, is described by 
\begin{equation}
    \mathbf{s} = \{\boldsymbol{x}_1, \cdots, \boldsymbol{x}_n\},\; \boldsymbol{x}_i \in \mathbb{R}^2
\end{equation} 
The start shape is denoted as $\mathbf{s}^s = \{\boldsymbol{x}^s_1, \cdots, \boldsymbol{x}^s_n\}$, and the goal is $\mathbf{s}^g=\{\boldsymbol{x}^g_1, \cdots, \boldsymbol{x}^g_n\}$.
The robot configuration $\mathbf{y} \in \mathcal{Y}$ encompasses the positions and orientations of the left and right end effectors (EEFs), with the orientation represented by the angle $\theta$ Specifically, $\theta$ denotes the yaw angle of each EEF.
\begin{equation}
    \mathbf{y} = \{ (\boldsymbol{x}_L, \theta_L), (\boldsymbol{x}_R, \theta_R)\} = \{(x, y, \theta)_L, (x, y, \theta)_R \}.
\end{equation}
The action $\mathbf{a} \in \mathcal{A}$ is the EEF movements,
\begin{equation}
    \mathbf{a} = \delta \mathbf{y} = \{ (\delta{x}, \delta{y}, \delta{\theta})_L, (\delta{x}, \delta{y}, \delta{\theta})_R \}.
\end{equation}
It is assumed that the obstacle geometry is given and convex~\cite{Guo2023}. The goal is to find a DLO deformation sequence $(\mathbf{s}^0, \cdots, \mathbf{s}^t, \cdots, \mathbf{s}^T)$ connecting $\mathbf{s}^s$ and $\mathbf{s}^g$, and the robot must execute appropriate actions that smoothly deform the DLO to track the intermediate shapes in obstacle-dense environments.

\section{Spatial Path Set Planning}
\label{sec:pathset_gen}
Given a DLO represented by a group of keypoints, a spatial \textit{path set} is the collection of keypoint paths, denoted by
\begin{equation}
    \begin{aligned}
        \mathcal{P} &= \{ \boldsymbol{p}_1, \cdots, \boldsymbol{p}_n \} \\
        \boldsymbol{p}_i &= (\boldsymbol{x}^0_i, \cdots, \boldsymbol{x}^m_i),\; m > 0
    \end{aligned}
\end{equation}
where $\boldsymbol{x}^j_i$ is the $j$-th waypoint of the $i$-th keypoint. The internal kinematic constraints of DLOs enforce that the path set forms a general class of \textit{homotopic} path set ~\cite{Jinghuang2023}. Intuitively, for $\boldsymbol{p}_i$ and $\boldsymbol{p}_j$ sharing identical initial and final positions, if one path can be continuously deformed to the other without crossing any obstacle, $\boldsymbol{p}_i$ and $\boldsymbol{p}_j$ are path homotopic.
This property enables efficient keypoint path planning: once the \textit{pivot path} $\boldsymbol{p}_p$ is established, other paths can be derived by deforming it. This centralized approach is more efficient, as path deformation is much lighter than planning each keypoint path individually. 
This section presents a systematic two-step method to generate a spatial path set for DLOs, including clearance-aware pivot path planning and passage-assisted path generation.
\subsection{Clearance-Aware Optimal Pivot Path Planning}
A pivot path is first planned with the RRT$^{*}$-based algorithm for a pivot point $\boldsymbol{x}_p$, which is selected as DLO's center. When manipulating a DLO in obstacle-dense environments, sufficient accessible free workspace along the path is essential to accommodate DLOs under significant deformation. We employ a clearance-aware cost metric for pivot path planning. The planning objective balances path length and obstacle clearance as following
\begin{equation}
    \ell(\boldsymbol{x}(\sigma)) = k_{\ell} Len(\boldsymbol{x}(\sigma)) + k_c \int^{\sigma}_{0} \frac{1}{\delta(\boldsymbol{x}(\sigma))}d\sigma,
\end{equation}
where $\sigma\in[0,1]$ is the path length parametrization and $\delta(\boldsymbol{x}(\sigma))$ is the obstacle clearance at node $\boldsymbol{x}(\sigma)$. $k_c$ and $k_{\ell}$ are positive weights, which offers a flexible way to encode suitable pivot paths tailored to different tasks. An example of pivot path planning with different parameters is shown in Fig. \ref{fig:345}(c). As expected, the clearance-aware pivot path (\textit{green path}) shows a larger accessible space for the DLO to deform and maintains a shorter path length.
\subsection{Passage-Assisted Path Set Generation}
This subsection further introduces a novel passage-assisted path deforming approach to generate paths for the DLO keypoints based on the pivot path.
\subsubsection{Passage Introduction}
Passages refer to regions between obstacles where free space shrinks dramatically~\cite{Sun2005,Guo2023}. 
Unlike prior works focused on passage detection or efficient sampling strategies in presence of narrow passages, our method primarily leverages passages as auxiliary geometric cues to facilitate the homotopic path set generation. Specifically, the shortest line segment between obstacles is used as a compact representation of the passage. We adopt the passage determination schema in~\cite{Jinghuang2024} to identify sparse and informative passages efficiently. A passage is considered valid if no obstacles intersect or fall within the circumcircle formed by the passage line segment, as illustrated in Fig. \ref{fig:345}(a). An example of passage construction for an obstacle-dense workspace is shown in Fig. \ref{fig:345}(b).

\subsubsection{Passage-Assisted Path Deforming}
\begin{figure*}
    \centering
    \includegraphics[width=0.92\textwidth]{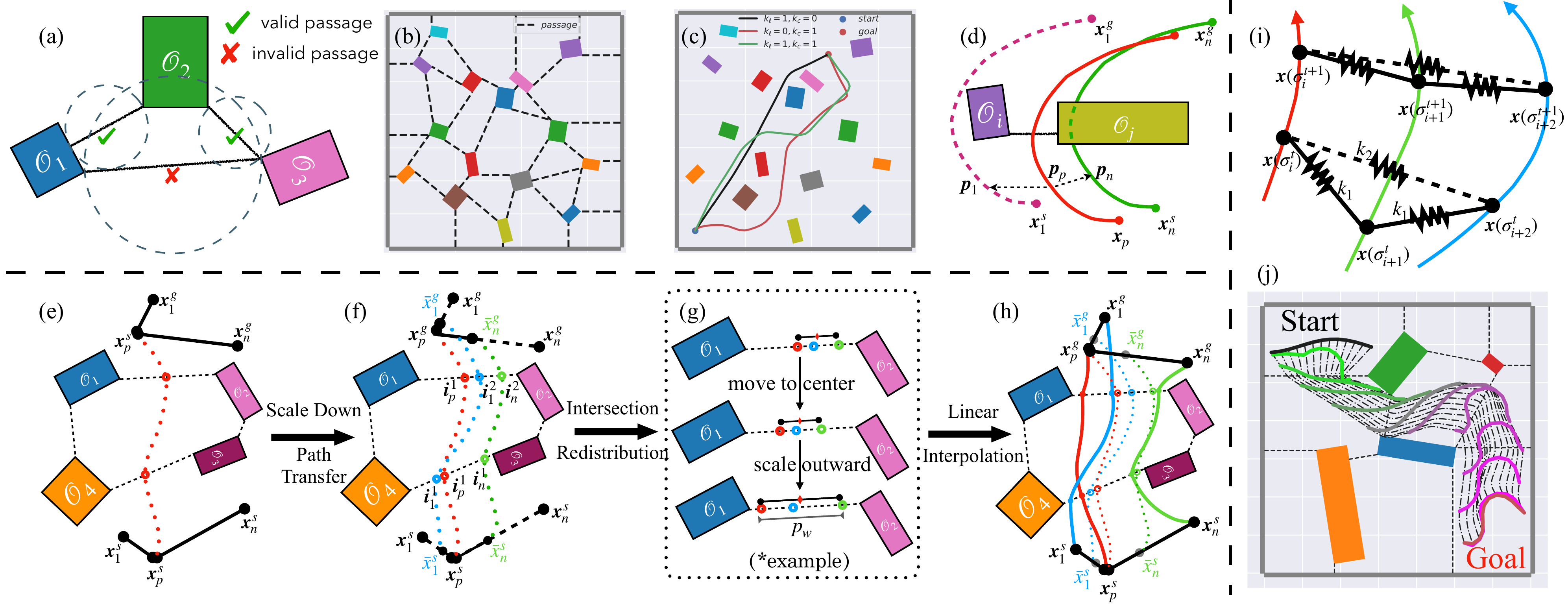}
    \caption{(a) Passage determination and validity checking routine. (b) Valid passages (dashed lines), workspace boundaries are treated as obstacles. (c) Pivot path planning with different parameters. (d) Failed case of pivot path transferring with linear interpolation. (e-h) Overview of passage-assisted homotopic path set generation. (i) Heuristic MS model along the path set. (j) Optimized deformation sequence in a constrained workspace with four obstacles.}
    \label{fig:345}
\end{figure*}

Given the start/target shapes and the optimal pivot path, directly transferring $\boldsymbol{p}_p$ by linear interpolation to other keypoints is an intuitive way. The transferred waypoint, $\boldsymbol{x}_i(\sigma)$, for the $i$-th keypoint is,
\begin{equation}\label{eq:linear_interpolation}
    \begin{aligned}
        &\boldsymbol{x}_i(\sigma) = \boldsymbol{x}_p(\sigma) + \Delta \boldsymbol{x}_i(\sigma)\\
        &\Delta \boldsymbol{x}_i(\sigma) = \Delta \boldsymbol{x}_i^s + \sigma (\Delta \boldsymbol{x}_i^g - \Delta \boldsymbol{x}_i^s) \\
        &\Delta \boldsymbol{x}_i^s = \boldsymbol{x}_i^s - \boldsymbol{x}_p^s, \; \Delta \boldsymbol{x}_i^g = \boldsymbol{x}_i^g - \boldsymbol{x}_p^g.
    \end{aligned}
\end{equation}
However, linear interpolation may not guarantee the homotopic property or free from collisions. For instance, as illustrated in Fig. \ref{fig:345}(d), direct pivot path transfer via linear interpolation causes the transferred path $\boldsymbol{p}_1$ no longer homotopic to $\boldsymbol{p}_p$ as they are separated by the obstacle $\mathcal{O}_i$. Additionally, $\boldsymbol{p}_2$ collides with the obstacle $\mathcal{O}_j$. Unlike \cite{Jinghuang2023}, which relies on hand-crafted rules to handle corner cases in linear interpolation, we propose a more general method that deforms the pivot path through shape scaling while explicitly leveraging available passages.

Figure~\ref{fig:345}(e-h) outlines a four-step workflow of path set generation. First, the passages traversed by the optimal pivot path are identified as in Fig.~\ref{fig:345}(e). 
Second, the start and goal shapes are uniformly scaled down about the pivot point $\boldsymbol{x}_p$ by a factor of $\gamma$. In general, $\gamma$ is chosen to be sufficiently small to ensure that all transferred paths remain collision-free when using the scaled shapes. Using the scaled start and goal shapes, path transfer through linear interpolation is applied to identify all intersections in the traversed passages, denoted as $(\boldsymbol{i}_1, \cdots, \boldsymbol{i}_n)$, as shown in Fig.~\ref{fig:345}(f). This step ensures that keypoint paths maintain the topological relationship. 
In the third step, the intersection points are redistributed along each passage to make efficient use of the passage width and avoid clustering as in Fig. \ref{fig:345}(g). In particular, for each traversed passage, the intersection points are first aligned to the passage center and then scaled outward along the passage. The maximum spacing between intersections is
\begin{equation}
    p_w = \min(p_d, \alpha L), \;0<\alpha <1,
\end{equation}
where $p_d$ is the passage width and $L$ is the DLO length. $\alpha$ is a hyperparameter selected to prevent overstretching. The final step connects the redistributed keypoints to the original start and goal shapes. This is achieved by transferring the raw keypoint paths (dashed lines) via linear interpolation between two successive traversed passages or between DLO shapes accordingly. Figure~\ref{fig:345}(h) presents an example of the path set that satisfies both spatial and topological constraints while maintaining large accessible free space along the path set.
\section{DLO Deformation Optimization}\label{sec:deform_opt}
Although the optimal spatial path set provides motion guidance for DLO manipulation by specifying the individual keypoint's path, it does not address how all keypoints should be coordinated to form a feasible DLO shape and move cohesively. To handle this problem, this section introduces the path set guided temporal DLO deformation optimization.

\subsection{Smooth Parametric Path Interpolation}
The discrete optimal spatial path set, $\mathcal{P}=\{\boldsymbol{p}_1,\cdots,\boldsymbol{p}_n\}$, is often insufficiently smooth. It is essential to query keypoint positions along the path in a differentiable manner. To achieve this, we first postprocess each path into a smooth and continuous curve. 
Given a single path, $\boldsymbol{p} = (\boldsymbol{x}^0, \cdots, \boldsymbol{x}^m)$, a weighted interpolation method is employed to achieve a smooth parametrization. 
For any given path parameter $\sigma$, the interpolated point $\boldsymbol{x}(\sigma)$ is computed as a weighted sum of the linear interpolations across all path segments. 
For the $k$-th segment of the path, the interpolated point $\boldsymbol{x}^k(\sigma)$ is given as
\begin{equation}
    \boldsymbol{x}^k(\sigma) = \boldsymbol{x}^k + (\boldsymbol{x}^{k+1}-\boldsymbol{x}^k)\frac{\sigma-\sigma^k}{\sigma^{k+1}-\sigma^k}
\end{equation}
A weight $w^k(\sigma)$ for each path segment $k$ is defined as,
\begin{equation}
    w^k({\sigma}) = \texttt{sigmoid}(\sigma-\sigma^k)\cdot \texttt{sigmoid}(\sigma^{k+1}-\sigma).
\end{equation}
where $\texttt{sigmoid}(x) = \frac{1}{1+e^{-ax}}$, and $a$ controls the transition sharpness. The interpolated position $\boldsymbol{x}(\sigma)$ are obtained by taking a weighted sum over all path segments,
\begin{equation}\label{eq:point_query}
    \boldsymbol{x}(\sigma) = \Gamma(\sigma, \boldsymbol{p}) = \frac{\sum^{m-1}_{k=0}w^k(\sigma)\boldsymbol{x}^k(\sigma)}{\sum^{m-1}_{k=0} w^k(\sigma)}
\end{equation}
A smooth and differentiable waypoint position can be generated via the path parameter $\sigma$, providing continuous gradients for subsequent deformation optimization.

\subsection{Deformation Optimization Problem Formulation}
With the optimal spatial path set $\mathcal{P}$, we synthesize deformation sequences via nonlinear optimization. The goal is to find a set of path parameters to organize the keypoints to form a feasible DLO shape in the following form
\begin{equation}
    \mathbf{s}^t = h(\mathcal{P}, \sigma_1^t, \cdots, \sigma_n^t),\; t = 0, \dots, T
\end{equation}
where $h()$ is a DLO shape mapping function and $t$ is the discrete step. The decision variable at the $t$-th step is a collection of the path parameters $\boldsymbol{\sigma}^t = [\sigma^t_1, \cdots, \sigma^t_n]^\top \in \mathbb{R}^{n}$.
Each $\boldsymbol{\sigma}^t$ encodes the progress of keypoints at $t$, guiding the deformation evolution. The full decision variable is $\boldsymbol{z} = [\boldsymbol{\sigma}^0,\cdots,\boldsymbol{\sigma}^{T}] \in \mathbb{R}^{T \times n}$, and the optimization problem is formulated as
\begin{equation}
    \begin{aligned}
        &\min_{\boldsymbol{z}}\; \mathcal{J}(\boldsymbol{z}; \mathcal{P})\\
        \mathrm{s.t.} \;\; &\mathbf{s}^t = h(\mathcal{P}, \boldsymbol{\sigma}^t)\\
        &\boldsymbol{lb}  \leq \boldsymbol{\sigma}^{t+1} - \boldsymbol{\sigma}^{t} \leq \boldsymbol{ub}\\
        &\boldsymbol{\sigma}^0 = \mathbf{0},\; \boldsymbol{\sigma}^T= \mathbf{1}
    \end{aligned}
\end{equation}
where $\mathcal{J}$ is the objective conditioned on the path set. $\boldsymbol{lb} $ and $\boldsymbol{ub}$ are the lower and upper bounds respectively, which control the smoothness of the deformation sequence.
\subsubsection{DLO Configuration Mapping}
The DLO shape $\mathbf{s}^t$ can be queried with $\mathbf{s}^t=h(\mathcal{P}, \boldsymbol{\sigma}^t)$, where $h()$ is inferred from the path set. Specifically, the $i$-th keypoint position at timestep $t$ is obtained from the planned path $\boldsymbol{p}_i$ from (\ref{eq:point_query})
\begin{equation}
    \boldsymbol{x}^t_i = \Gamma(\boldsymbol{p}_i, \sigma^t_i), i=1,\cdots,n
\end{equation}
The DLO candidate shape is then given by
\begin{equation}
    {\mathbf{s}}^t = h(\mathcal{P}, \boldsymbol{\sigma}^t) = [\Gamma(\boldsymbol{p}_1, \sigma^t_1), \cdots,  \Gamma(\boldsymbol{p}_n, \sigma^t_n)].
\end{equation}
\subsubsection{Objective With Mass-Spring Model}
In DLO manipulation, one intuitive objective is to keep DLO in smooth shapes with low potential energies, avoiding overstretching and overcompression. As shown in Fig. \ref{fig:345}(i), we employ a heuristic mass-spring (MS) model~\cite{dom_model} to obtain the accumulated potential energy. 
For MS model, the potential energy $E(\mathbf{s}^t)$ can be calculated using Hooke’s Law,
\begin{equation}
    \begin{aligned}
        E(\mathbf{s}^t) &= \frac{1}{2}\sum^{n-1}_{i=1} k_1(||\boldsymbol{x}^t_{i+1}-\boldsymbol{x}^t_i||_2 - L_{i+1,i})^2 \\
        &+ \frac{1}{2}\sum^{n-2}_{j=1}k_2(||\boldsymbol{x}^t_{j+2}-\boldsymbol{x}^t_j||_2 - L_{j+2,j})^2
    \end{aligned}
\end{equation}
$L_{i+1,i}$ is the rest length between $i$-th and $i+1$-th points. $k_1$ and $k_2$ are virtual spring stiffness for MS model. Though the parameter identification for $k_1$ and $k_2$ is non-trivial~\cite{adaptigraph}, the deformation sequence serves as a reference and is not required to be physically aligned with the DLO. We simplify parameter settings as in Table \ref{table:hyp-param}. The overall objective is 
\begin{equation}
    \mathcal{J}(\boldsymbol{z}; \mathcal{P}) = \sum^T_{t=0} E(\mathbf{s}^t).
\end{equation}
The deformation optimization problem is fully defined and can be solved offline. An example of the path-set-guided deformation optimization result is illustrated in Fig.~\ref{fig:345}(j). The results demonstrate that our approach efficiently generates smooth and feasible DLO configurations. {Overall, the complexity of the proposed framework scales linearly with the number of keypoints $n$ and discrete resolution $T$. This linear scaling makes the approach suitable for DLOs with a moderate to large number of keypoints.}
\section{Neural MPC Tracking}
Given an optimal deformation sequence, the next challenge is to manipulate DLOs to accurately track reference shapes in constrained environments. {This study employs a neural model predictive control (MPC) tracking controller, framing deformation tracking as a finite-horizon optimization problem, which is solved using gradients from a learned deformation model. This design enables the deformation dynamics to be learned from data through a neural approximation, facilitating smooth generalization to novel DLOs. This section outlines the learning process for the DLO deformation model and provides a comprehensive overview of the neural MPC tracker.}
\subsection{Deformation Model Learning}
To predict the complex interactions between the EEFs and the DLO,
\begin{figure}[t]
    \centering
    \includegraphics[width=0.8\linewidth]{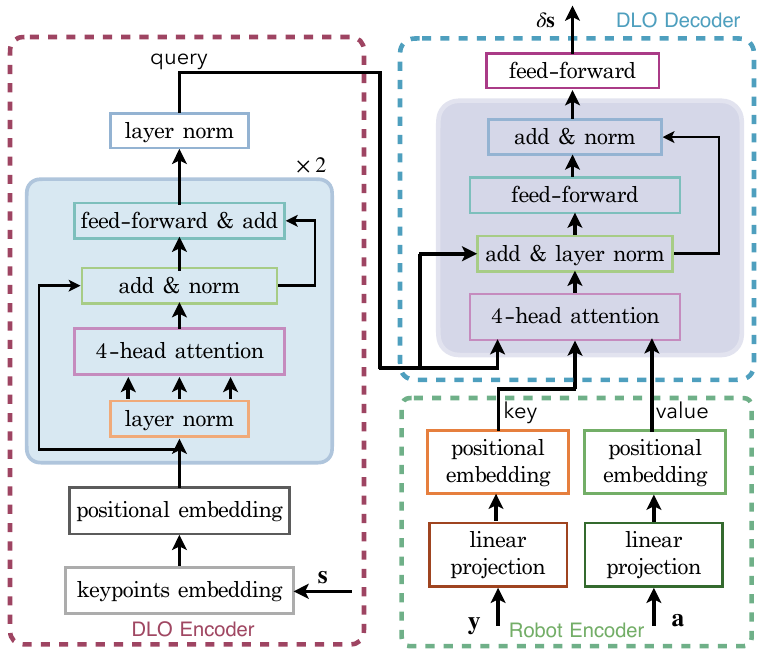}
    \caption{Deformation model architecture with three components: DLO encoder, robot encoder, and DLO decoder.}
    \label{fig:gdm_framework}
\end{figure}
a novel model architecture is proposed to learn a discrete transition function $\boldsymbol{f}: \mathcal{S} \times  \mathcal{Y} \times \mathcal{A} \rightarrow \delta \mathcal{S}$. The deformation model takes as input the current state-action pair and predicts the shape change, which can be written as
\begin{equation}
    \delta{\mathbf{s}} = \boldsymbol{f}(\mathbf{s}, \mathbf{y}, \mathbf{a}).
\end{equation} 
\subsubsection{Model Design}
In this work, a Transformer-based deformation model (T-DM) is proposed as shown in Fig. \ref{fig:gdm_framework}. Specifically, the self-attention mechanism in the DLO encoder module enables the model to encode the internal relationships between the DLO keypoints. This is crucial because the DLO deformation is influenced by both local interactions (e.g., tension between adjacent points) and global dependencies (e.g., the overall shape and boundary conditions). Meanwhile, the cross-attention module in the DLO decoder facilitates learning the interaction between the EEF motions and the resulting DLO deformation. DLO keypoints serve as the query, EEF poses as the key, and EEF motions as the value in multi-head cross attention.
This design can effectively capture how robot actions propagate through DLO, offering coherent deformation predictions.
\subsubsection{Model Training}
In offline data collection, EEFs are moved to random target poses in unconstrained environments. The simulation dataset has about $50$k transitions and the deformation model is trained by minimizing $\mathcal{L}$,
\begin{equation}\label{eq: global_FKM}
    \min_{\boldsymbol{\theta}}\; \mathcal{L}=\frac{1}{B}\sum^{B}_{k=1} ||\delta \mathbf{s}_k - \boldsymbol{f}_{\boldsymbol{\theta}}(\mathbf{s}_k,\mathbf{y}_k, \mathbf{a}_k)||_2^2.
\end{equation}
$\boldsymbol{f}_{\boldsymbol{\theta}}$ is trained with AdamW optimizer. {To reduce computational load, the DLO and robot encoders are kept frozen and only the decoder is updated in online adaptation. The frozen encoders provide stable latent representations, while the decoder adapts to distribution shifts by fine-tuning a small subset of parameters (e.g., cross-attention) using online interaction data. This lightweight design enables efficient online adaptation with minimal computation and reduced overfitting risk.}
\subsection{Neural MPC Design}
Given a well-trained deformation model, the constrained deformation tracking can be formulated as a finite-horizon optimization problem and solved in a MPC framework. MPC optimizes a sequence of actions by solving following problem
\begin{equation}\label{nmpc}
\begin{aligned}
    \mathbf{a}^{*}_{0:H-1}&=\arg\min_{\mathbf{a}_{0:H-1}} c_t(\mathbf{s}_{H}, \mathbf{y}_{H}) + \sum_{k=0}^{H-1} c(\mathbf{s}_{k}, \mathbf{y}_{k}, \mathbf{a}_{k}) \\
    \mathrm{s.t.} \quad &\mathbf{s}_{k+1} - \mathbf{s}_{k} - \boldsymbol{f}_{\boldsymbol{\theta}}(\mathbf{s}_k, \mathbf{y}_k, \mathbf{a}_k) =\mathbf{0}\\
    &\mathbf{y}_{k+1} - \mathbf{y}_k - \mathbf{a}_k =\mathbf{0},\;\mathbf{s}_0-\mathbf{s}_{\textcolor{black}{curr}}=\mathbf{0}\\
    & \mathbf{a}_{min} \leq \mathbf{a}_k \leq \mathbf{a}_{max}
\end{aligned}
\end{equation}
where $H$ is the prediction horizon and $\mathbf{s}_{curr}$ is the current shape. The deformation predictions are generated by rolling out the learned model. $c$ and $c_t$ represent the path and terminal objectives, respectively. 
The objectives are designed as:
\subsubsection{Deformation Tracking}
The tracking objective minimizes the errors between the current and reference shapes,
\begin{equation}
    c_{track}(\mathbf{s}) = (\mathbf{s}-\mathbf{s}_{ref})^{\top}\mathbf{Q}(\mathbf{s}-\mathbf{s}_{ref}) / 2,
\end{equation}
where $\mathbf{Q} \succeq 0$ is a semi-positive definite weight matrix. $\mathbf{s}_{ref}$ is obtained from the optimized deformation sequence. 
\subsubsection{Obstacle Avoiding}
The obstacle avoidance objective penalizes the proximity of keypoints and EEFs to obstacles. The obstacle-avoidance objective, $c_{obs,i}(\mathbf{s}, \mathbf{y})$, between the $i$-th keypoint and the closet obstacle, $\mathcal{O}$, is defined as
\begin{equation}
    {c_{obs,i}(\mathbf{s})}= {1/{\phi^2_{\mathcal{O}}(\boldsymbol{x}_i)}}, \text{ if } \phi_{\mathcal{O}}(\boldsymbol{x}_i) \leq d \text{ else } 0
\end{equation}
where $d$ is a distance threshold. $\phi_{\mathcal{O}}(\boldsymbol{x}_i)$ is the signed distance value between the $i$-th keypoint and the obstacle. For the EEFs' obstacle avoidance, the objective is similarly as
\begin{equation}
    c_{obs}(\mathbf{y}) = 
    \begin{cases}
        {0}, &{\phi_{\mathcal{O}}(\boldsymbol{x}_{R|L}) > d} \\
        {1/{\phi_{\mathcal{O}}^2(\boldsymbol{x}_L)} + 1/{\phi_{\mathcal{O}}^2(\boldsymbol{x}_R)}}, & \color{black}{\phi_{\mathcal{O}}(\boldsymbol{x}_{R|L}) \leq d}
    \end{cases}
\end{equation}
where $\boldsymbol{x}_L$ and $\boldsymbol{x}_R$ denote the positions of the right and left EEF, respectively.
\begin{table}[t]
	\centering
	\caption{Key Hyperparameters}
	\begin{tabular}{c c}
    \toprule
    {\textbf{Simulator Settings}} &{\textbf{Value(s)}}\\
    \midrule
    {Cable size} &{length: $L=0.3$ m, radius: $0.02$ m}\\
    {Joint stiffness \& damping} &{$0.02$ Nm/rad, $0.03$ Nms/rad}\\
    \toprule
    {\textbf{Deformation Planning}} &{\textbf{Value(s)}}\\
    \midrule
    {Pivot path weights} &{$k_c$= 1, $k_{\ell}$= 1}\\
    {Virtual stiffness} &{$k_1= 2$ N/m, $k_2= 1$ N/m}\\
    {Discrete timestep} &{$T=50$}\\
    {Lower bound} &{$\boldsymbol{lb}=[0.01,\cdots,0.01]^\top$} \\
    {Upper bound} &{$\boldsymbol{ub}=[0.1,\cdots,0.1]^\top$}\\
    \toprule
    {\textbf{Deformation Tracking}} &{\textbf{Value(s)}}\\
    \midrule
    {Distance threshold} &{$d=0.005$ m}\\
    {MPC horizon} &{$H=5$}\\
    {MPC max. iter.} &{$15$}\\
    {Objective weight parameters} &{$\lambda_1=150$, $\lambda_2=0.01$, $\lambda_3=1$}\\
    \bottomrule
    \end{tabular}
    \label{table:hyp-param}
\end{table}
The overall obstacle avoidance objective is
\begin{equation}
    c_{obs}(\mathbf{s}, \mathbf{y}) = \sum^n_{i}c_{obs,i}(\mathbf{s}) + c_{obs}(\mathbf{y})
\end{equation}
\subsubsection{Control Regularization}
The control objective minimizes the movements of EEFs as following
\begin{equation}
    c_{ctrl}(\mathbf{a}) = \mathbf{a}^{\top}\mathbf{R}\mathbf{a}/2,
\end{equation}
where $\mathbf{R} \succeq 0$ is a semi-positive definite weight matrix. The overall objective combines multiple objectives as 
\begin{equation}
    c(\mathbf{s}, \mathbf{y}, \mathbf{a}) = \lambda_1 c_{track} + \lambda_2 c_{obs} + \lambda_3 c_{ctrl}
\end{equation}
where $\lambda_1$, $\lambda_2$ and $\lambda_3$ are positive weight hyperparameters for objective terms. These hyperparameters are selected to ensure that the problem is well-posed. The terminal objective $c_t$ only includes $c_{track}$ and $c_{obs}$. As $\boldsymbol{f}$ is differentiable, a gradient-based optimizer can be utilized to solve the problem efficiently.
\section{Simulation Experiments}
We conduct simulations to validate the proposed framework in MuJoCo, with deformable cables modelled as rigid links connected by hinge joints to efficiently simulate elastic deformation.
A $0.3$ m cable with 13 uniformly distributed keypoints is rigidly grasped at both ends in a $0.7 \times 0.7$ m workspace. The pivot path planner uses RRT$^{*}$ from OMPL~\cite{ompl} with collisions handled by FCL~\cite{fcl}. Deformation optimization is solved using IPOPT~\cite{ipopt}, leveraging JAX~\cite{jax} for automatic differentiation and acceleration. Experiments were conducted on a desktop PC with Intel Core i9-13900K and RTX 4070.
\subsection{Deformation Planning Results}
\begin{figure}[tb]
    \centering
    \includegraphics[width=0.48\textwidth]{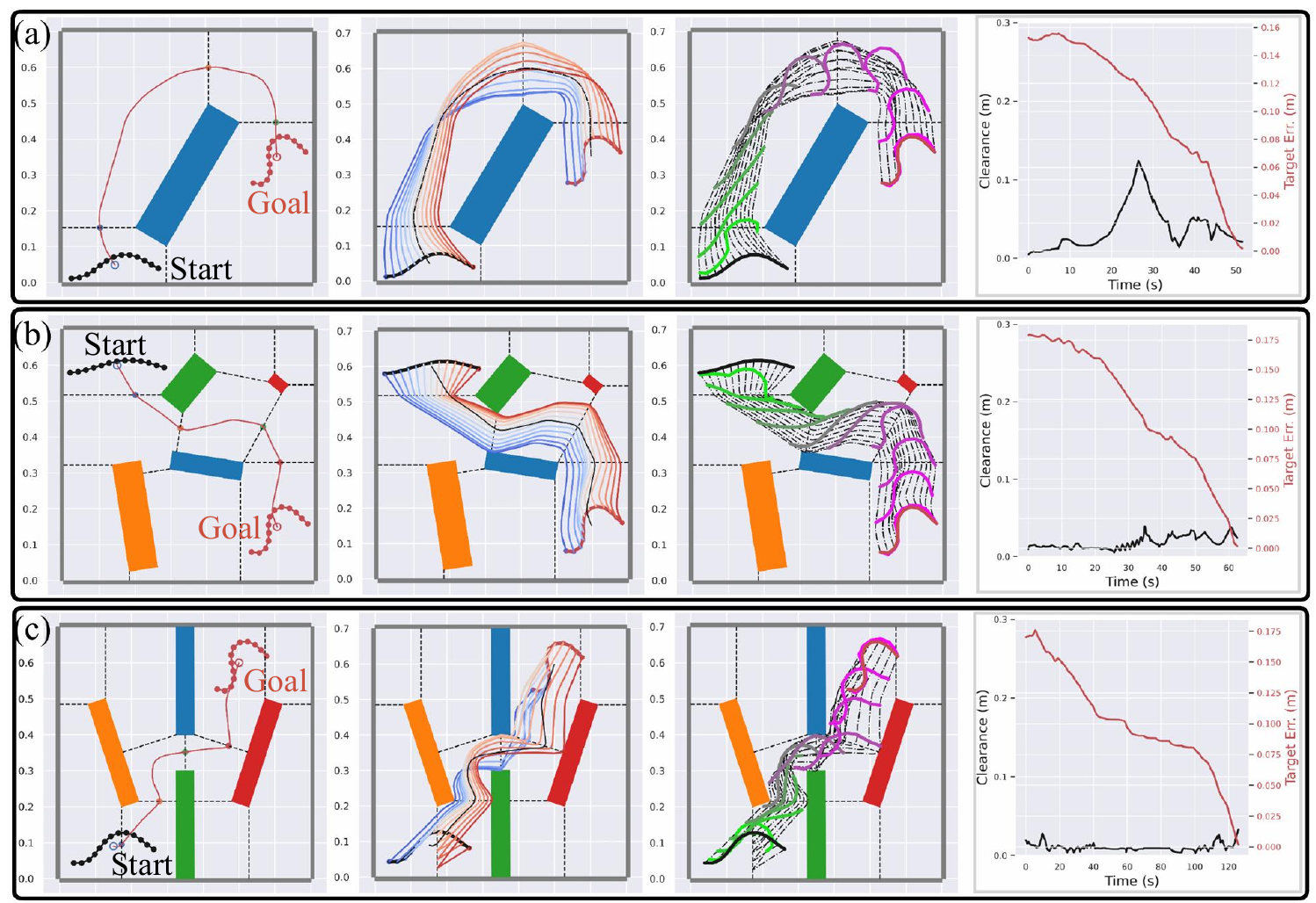}
    \caption{Deformation planning and tracking results. (a) single obstacle avoidance, (b) multiple obstacle avoidance, and (c) maze navigation tasks.}
    \label{fig:planning_res}
\end{figure}

Each subplot in Fig.~\ref{fig:planning_res} represents a specific environment case, showcasing the pivot path planning, spatial path set, optimized deformation sequence, and neural tracking performance in sequence. The first three columns highlight the deformation planning results, providing a concise overview of the process.
\begin{figure}[tb]
    \centering
    \includegraphics[width=0.98\linewidth]{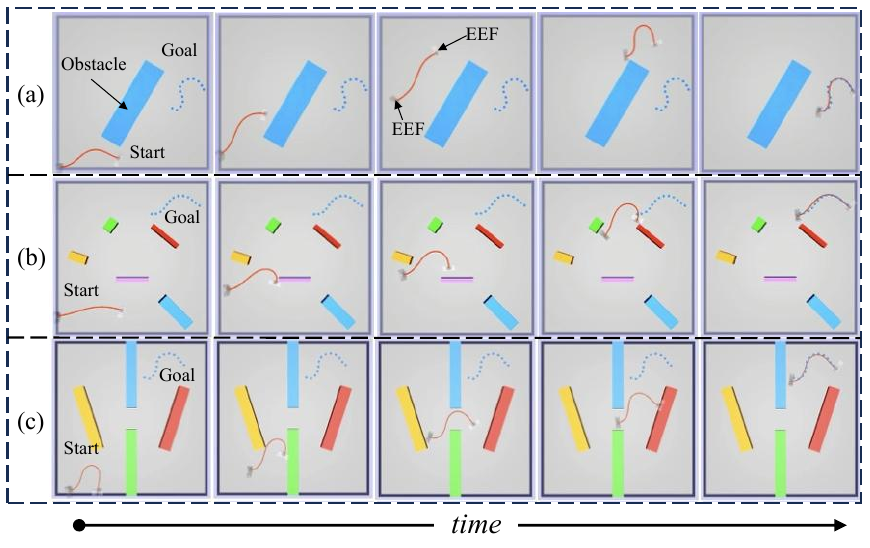}
    \caption{Manipulation snapshots of: (a) single obstacle avoidance, (b) multiple obstacle avoidance, (c) maze navigation task.}
    \label{fig:so_mo_and_ma_snapshots}
\end{figure}
In pivot path planning, the clearance-aware optimal pivot paths in the first column demonstrate a wide accessible free workspace along the trajectories. Notably, the pivot path in Fig. \ref{fig:planning_res}(a) approaches the target from above the obstacle, prioritizing sufficient manipulation space.
Spatial path sets in the second column, generated by the passage-assisted path set generation, maintain homotopic kinematic constraint. Each spatial path set utilizes passage widths effectively and offers motion guidance for DLO manipulation. 
The optimized DLO shapes are visualized in the last column of Fig. \ref{fig:planning_res}, showing smooth deformation along the planned path set. In particular, the cable goes straight while passing through the second passage before bending back to align with the target shape. The proposed deformation planner shows robust performance in various constrained environments.
\subsection{DLO Manipulation Results}
Representative simulation tasks of increasing complexity are designed to evaluate the framework, including \textit{single obstacle avoidance}, \textit{multiple obstacle avoidance}, and \textit{maze navigation} tasks. 
Two metrics are reported in deformation tracking: the minimum clearance between the DLO and obstacles, and the shape error relative to the target. The two metrics are given in the last column of Fig. \ref{fig:planning_res}.
\subsubsection{Single Obstacle Avoidance} In this task, the workspace features a single obstacle. 
We report two trials of single obstacle avoidance manipulation and the snapshots are visualized in Fig. \ref{fig:so_mo_and_ma_snapshots}(a). For example, in the second environment of Fig. \ref{fig:so_mo_and_ma_snapshots}(a), the target shape was occluded by a square obstacle ($0.4$ m $\times$ $0.1$ m) centered at $(0.35, 0.3)$ m. The deformation tracking performance is given in the fourth column of Fig. \ref{fig:planning_res}(a). The consistently positive clearance confirms that no collisions occurred during manipulation and the final target error was below $2$ cm.
\subsubsection{Multiple Obstacle Avoidance} This task involves a workspace populated with multiple obstacles randomly arranged, and the goal is positioned far from the start configuration. 
As shown in Fig. \ref{fig:planning_res}(b), the shape error continuously decreased throughout the process, achieving a final average target error less than $1$ cm, while maintaining positive obstacle clearance. Snapshots of this manipulation are illustrated in the second row of Fig. \ref{fig:so_mo_and_ma_snapshots}(b). Additionally, the manipulation processes for another multiple-obstacle avoidance task is presented in the first row of Fig. \ref{fig:so_mo_and_ma_snapshots}(b). In these scenarios, the EEFs skillfully deformed the cable through the narrow gaps between obstacles. 
\subsubsection{Maze Navigation} 
We test the proposed method by guiding DLOs through constrained maze environments featuring narrow, corridor-structured passages, where only a single feasible path is available.
Figure \ref{fig:so_mo_and_ma_snapshots}(c) showcases the snapshots from manipulation examples in two different maze configurations. 
For the first maze task, the corresponding deformation tracking performance is depicted in Fig. \ref{fig:planning_res}(c). The plot illustrates that the target error is minimized sufficiently ($<2$ cm) to match the target shape, demonstrating that the neural MPC can accurately track the planned deformation sequence. In addition, obstacle clearance remains positive, ensuring safe navigation through the maze. These results validate the robustness of the proposed framework in navigating maze-like environments.
\section{Real-world Experiments}

\subsection{Hardware Setup}
The hardware setup in Fig.~\ref{fig:cables}(a) consists of two UR5e robots manipulating cables on a tabletop, with each arm holding one cable tip. Thirteen uniformly spaced points along the cable were tracked at $30$ Hz using Lucas–Kanade algorithm with a RealSense RGB-D camera. The neural MPC tracker generates real-time control commands at $10$ Hz. Cable details are provided in Table~\ref{table:rw_dlo}.
\begin{table}[tb]
	\centering
	\caption{Real-world DLO Properties}
    \begin{tabular}{ccccc}
    \toprule
    Cable Type & $L$ ($m$) & $d$ ($mm$)  & $E$ (GPa)  & \textbf{$\mu$} ($g/m$) \\
    \midrule
     Cotton & 0.22 & 7.5  & 0.37$\sim$0.49 & 17.227\\
     Paracord  & 0.34, 0.17& 5 & 0.2$\sim$0.5 & 37.471 \\
     Rubber  & 0.34, 0.17 & 9 & (0.1$\sim$1)$\times10^{-4}$ & 38.765 \\
     Ethernet  & 0.34 & 5 & 10$\sim$18 & 47.967 \\
     USB& 0.34 & 8 & 25$\sim$30 & 91.824 \\
     HDMI& 0.25 & 5.5 & 5$\sim$10 & 42.818 \\
     Silicone& 0.27& 11.5 & (3$\sim$8)$\times10^{-3}$ & 79.548\\
    \bottomrule
    \end{tabular}
    \begin{tablenotes}
        \centering
        \item $d$: diameter; $E$: Young's module; $\mu$: linear density.
    \end{tablenotes}
    \label{table:rw_dlo}
\end{table} 
\subsection{Deformation Model Performance}
The test objects include Ethernet, Paracord, rubber, and USB cables. For each cable, $\sim$25k transitions were collected with random EEF motions and teleoperation~\cite{dvrk}. The deformation model, pretrained on simulation data, is fine-tuned using real-world data. We compare its performance against two sequence modeling baselines: (1) a one-dimensional convolution (Conv1D) architecture based DM (C-DM) and (2) a Bi-LSTM architecture based DM (B-DM), 
\begin{figure}[tb]
    \centering
    \includegraphics[width=0.9\linewidth]{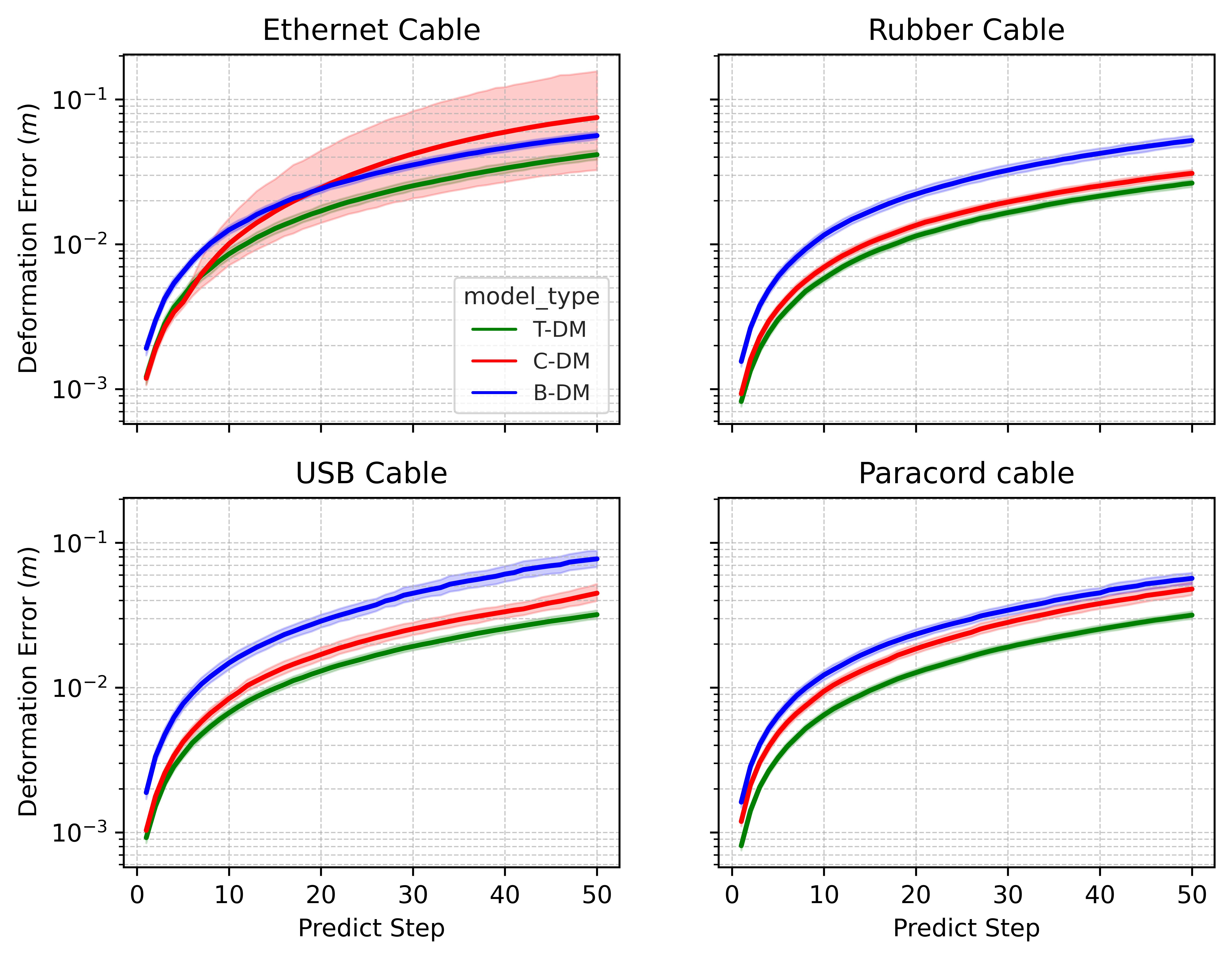}
    \caption{Multi-step deformation prediction comparisons on different cables.}
    \label{fig:deformation_comp}
\end{figure}both designed to approximate deformation models. {These two baselines are representative sequence modeling paradigms in prior deformable object dynamics research. The convolution neural network captures local dependencies through convolutional filters, offering computational efficiency~\cite{Zhang2025}. Bi-LSTM model captures bidirectional temporal correlations through recurrent connections, making it a strong baseline for long-term sequence prediction~\cite{Yang2021}. Comparing with these widely adopted architectures allows us to validate the T-DM's advantages in modeling global dependencies and complex spatial relationships essential for accurate deformation prediction.}
We evaluate prediction accuracy with 
$||\hat{\mathbf{s}}_{t+p} - \mathbf{s}_{t+p}||_2$, measuring the difference between the predicted and actual shapes at horizon $p$. Figure~\ref{fig:deformation_comp} shows that T-DM consistently achieves lower errors for both short and long-horizon predictions. Specifically, on USB cable, T-DM reduces the average long-horizon prediction error by $54.7\%$ compared to C-DM, and by $70.4\%$ compared to B-DM. The B-DM model performs the worst, indicating that B-DM struggles in capturing long-range dependencies. Though C-DM shows comparable results to T-DM on the rubber cable, it suffers from significantly higher prediction errors (up to $45$\%) on the other three cables. This variation underscores that C-DM is effective in capturing local spatial features, but struggles to model critical global dependencies. T-DM models long-range interactions between DLO and EEFs, enhancing generalization and accuracy across various DLOs.
\subsection{DLO Manipulation Results}
This section evaluates the proposed approach across diverse DLO materials and constrained environments.
\subsubsection{Unconstrained Deformation Control}
\begin{figure*}
    \centering
    \subfloat[]{\includegraphics[width=0.485\linewidth]{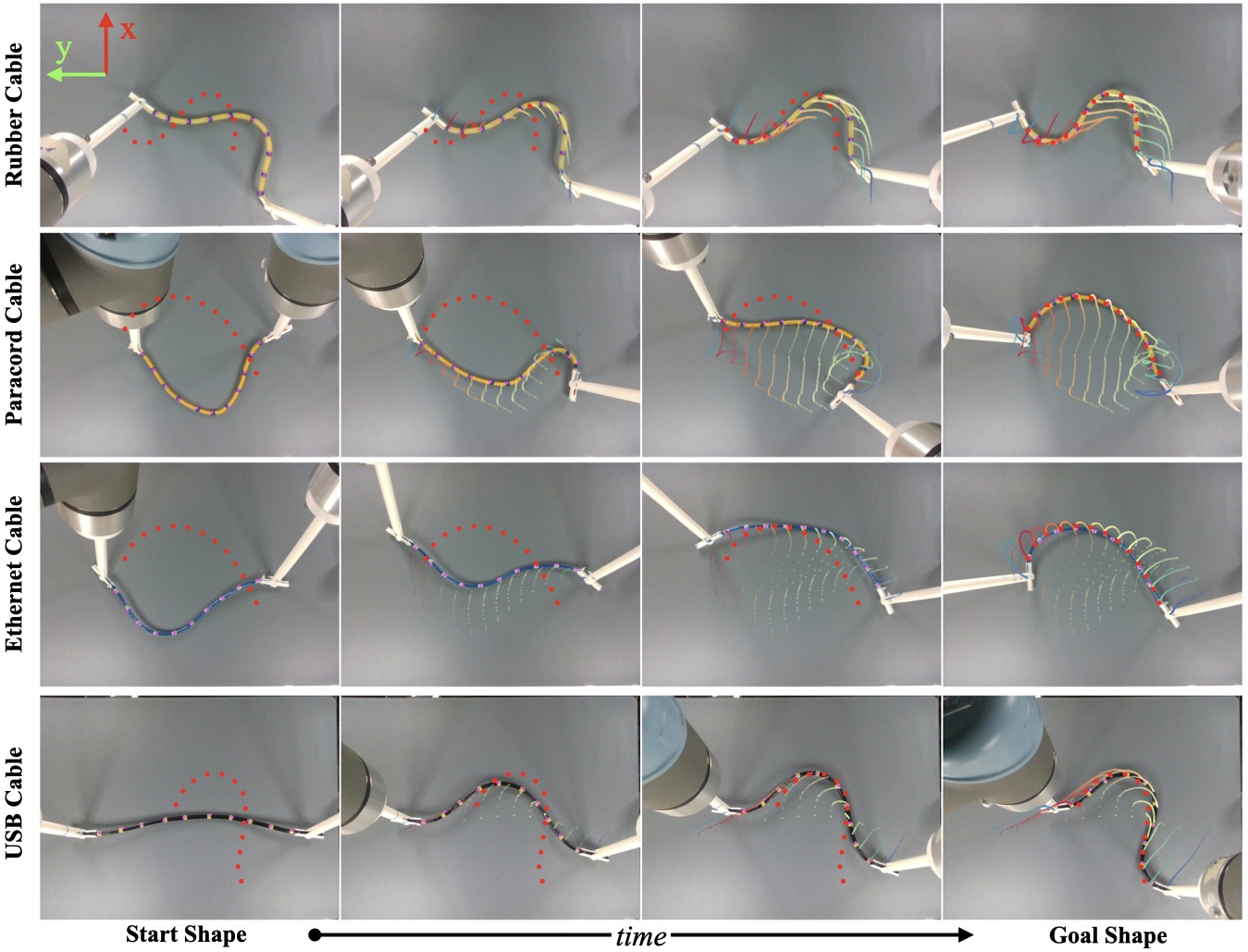}}%
    \hfill
    \subfloat[]{\includegraphics[width=0.485\linewidth]{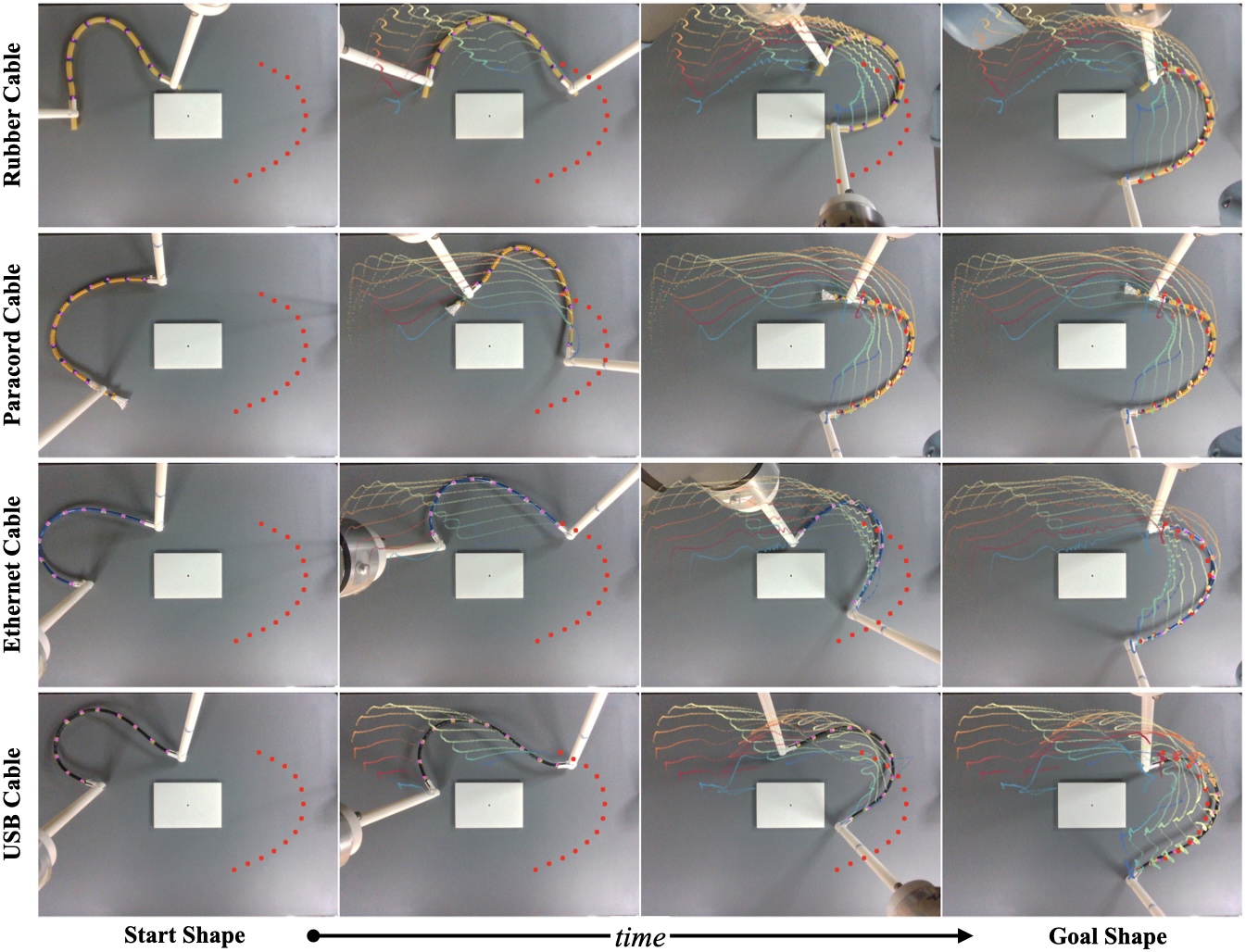}}%
    \caption{The target shapes are indicated by red points and the colored lines represent the cable keypoint trajectories. (a) Manipulation snapshots of unconstrained deformation control. (b) Manipulation snapshots of single obstacle avoidance tasks.}
    \label{fig:real_static_and_so_snapshots}
\end{figure*}

The proposed T-DM MPC controller is firstly validated on unconstrained shape formation tasks. As illustrated in Fig. \ref{fig:real_static_and_so_snapshots}(a), the cable will be deformed into different target shapes. To assess generalizability, the start and goal shapes are generated online, and and tracking error is measured by the $L2$ norm to the target.
\begin{table}[tb]
	\centering
	\caption{Deformation Control Results (with $H=5$)}
	\begin{tabular}{c c | c c}
    \toprule
    {Cable} &{Target Error (cm)} &{Cable} &{Target Error (cm)}\\
    \midrule
    {Paracord} &{$1.95 \pm 0.9$ } &{Ethernet} &{$1.69 \pm 0.8$ }\\
    {Rubber} &{$1.56 \pm 0.8$ } &{USB} &{$1.71 \pm 0.9$ } \\
    \bottomrule
    \end{tabular}
    \label{table:static_deform_result}
\end{table}
\begin{table}[tb]
    \centering
    \caption{Deformation Control Comparison }
    \begin{tabular}{cc|cc}
    \toprule
    Method     &  Target Error (cm) & Method & Target Error (cm)\\
    \midrule
    Jacob-IK  &  $2.56 \pm 1.53$ &C-DM MPC   & $1.95 \pm 0.91$\\
    B-DM MPC &  $2.12 \pm 0.73$ &
    {T-DM MPC} &  $\textbf{1.70} \pm \textbf{0.86}$\\
    \bottomrule
    \end{tabular}
    \label{tab:dt_comp}
\end{table}
Figure~\ref{fig:real_static_and_so_snapshots}(a) shows the manipulation snapshots. The target error steadily decreased below $2$ cm, demonstrating successful cable deformation. Table~\ref{table:static_deform_result} summarizes the statistical results from 10 experiment trials for each cable. These tasks evaluated the neural controller's precise deformation control, including smooth curves and significant bidirectional bends with seamless transitions. Notably, the neural MPC maintained consistent performance on all tested cables, despite significant variations of physical properties. {We compare the proposed T-DM based MPC with C-DM and Bi-LSTM based MPC methods, as well as a standard Jacobian-based inverse kinematic controller (Jaco-IK)~\cite{David2016}. The results, summarized in Table~\ref{tab:dt_comp}, show that our method consistently outperforms the other approaches across all the cables.}

\subsubsection{Single Obstacle Avoidance} As illustrated in Fig.~\ref{fig:real_static_and_so_snapshots}(b), the cable should be deformed around a rectangular obstacle to the target on the opposite side. A $0.08$ m $\times$ $0.12$ m obstacle is located at the workspace center. Each cable started from a slightly different start configuration, reflecting natural variations in placement. Despite these differences, all cables followed the same planned deformation trajectory from the hierarchical deformation planner. The neural tracker successfully adapted to the varying properties of each cable, exhibiting consistent motion execution across trials.
All cables were deformed to the targets without collisions. The average final target error across all cables was $1.77$ cm, with individual errors ranging from $1.01$ cm (Paracord cable) to $2.86$ cm (USB cable). Additionally, the minimum obstacle distance throughout the manipulation remained positive, with a mean obstacle clearance of $1.6$ cm, indicating safe manipulations. 
\begin{figure}[t]
    \centering
    \includegraphics[width=0.9\linewidth]{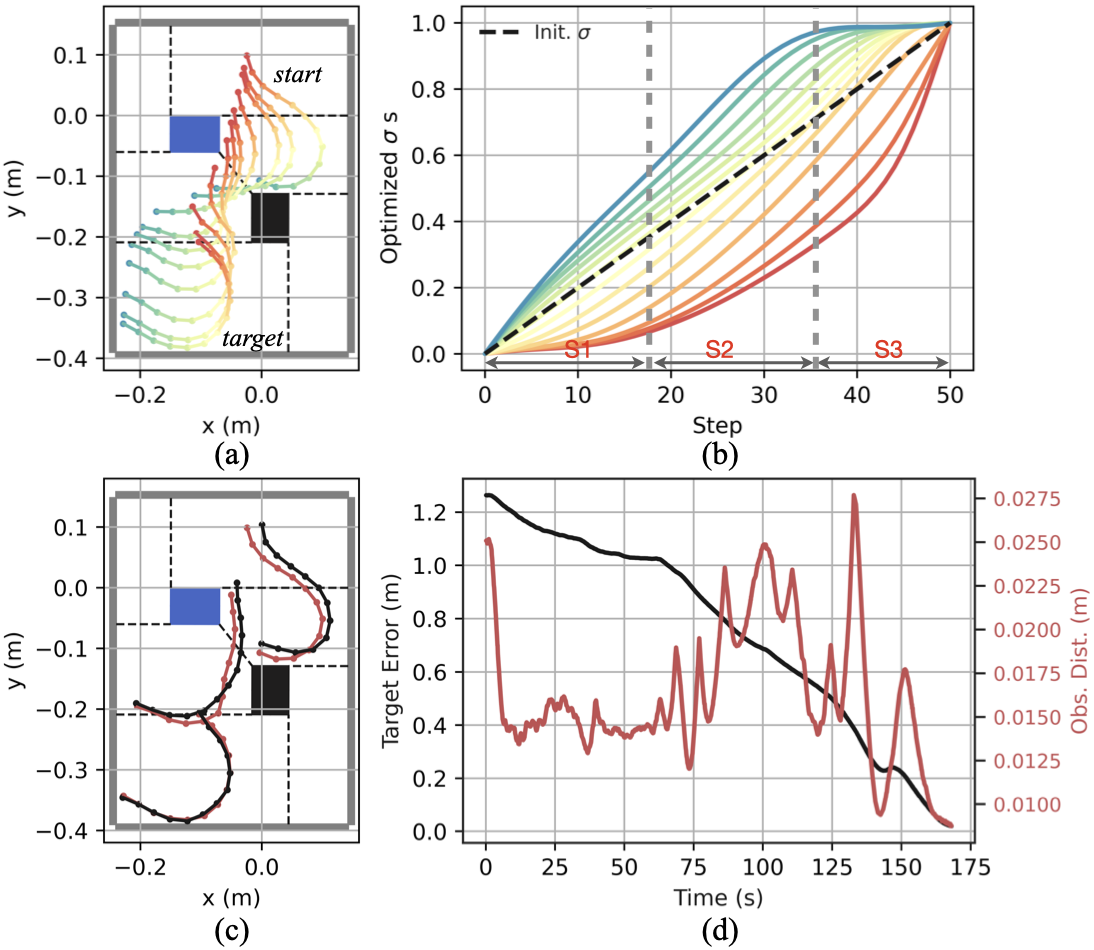}
    \caption{(a) Optimized global deformation flow. (b) Optimized decision variables $\boldsymbol{z}$. (c) Deformation tracking: red for reference shapes, black for actual shapes. (d) Target error and cable-to-obstacle distance profiles.}
    \label{fig:opt_res}
\end{figure}
The manipulation snapshots are visualized in Fig. \ref{fig:real_static_and_so_snapshots}(b).
\begin{figure*}[th]
    \centering
    \subfloat[]{\includegraphics[width=0.485\linewidth]{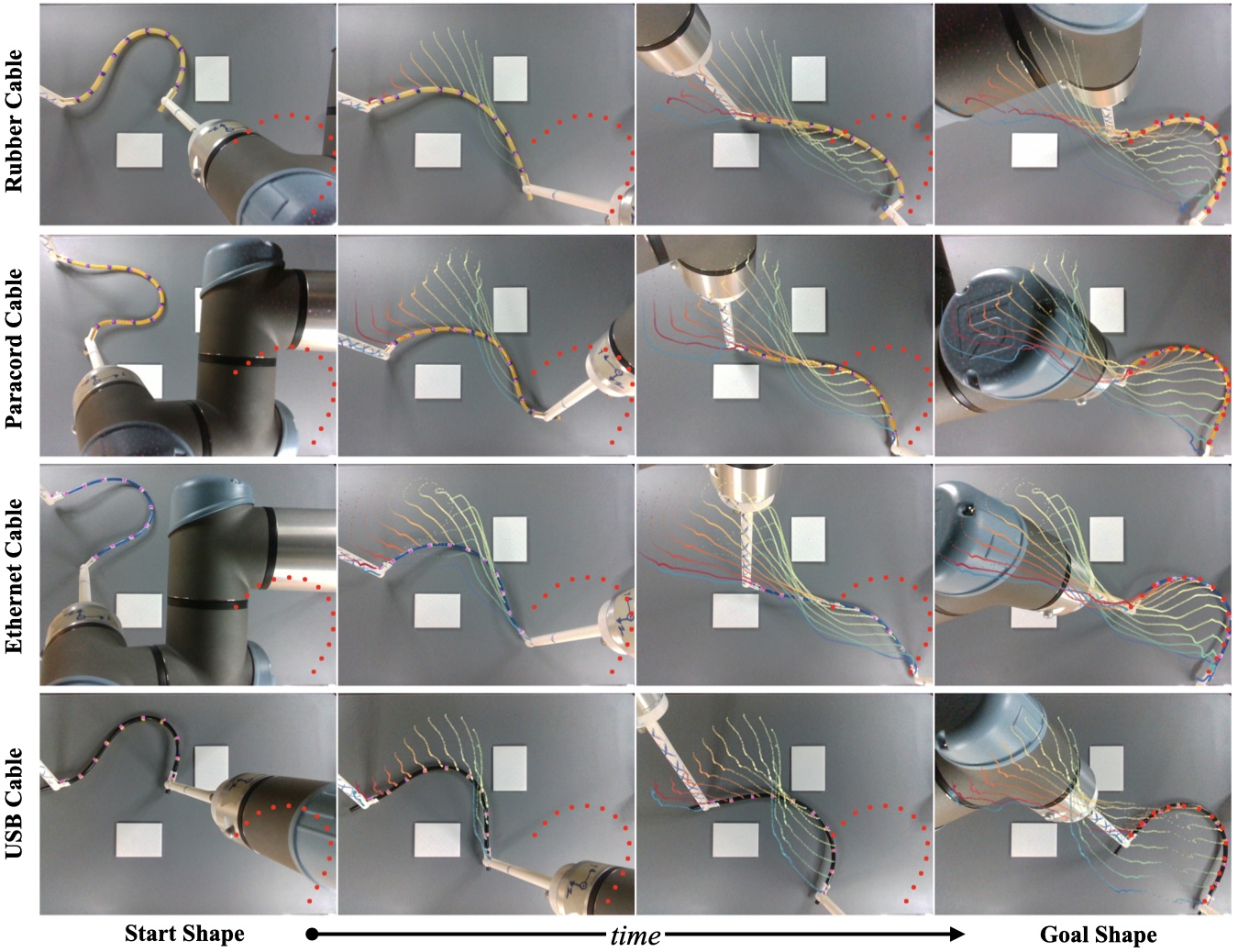}}%
    \hfil
    \subfloat[]{\includegraphics[width=0.485\linewidth]{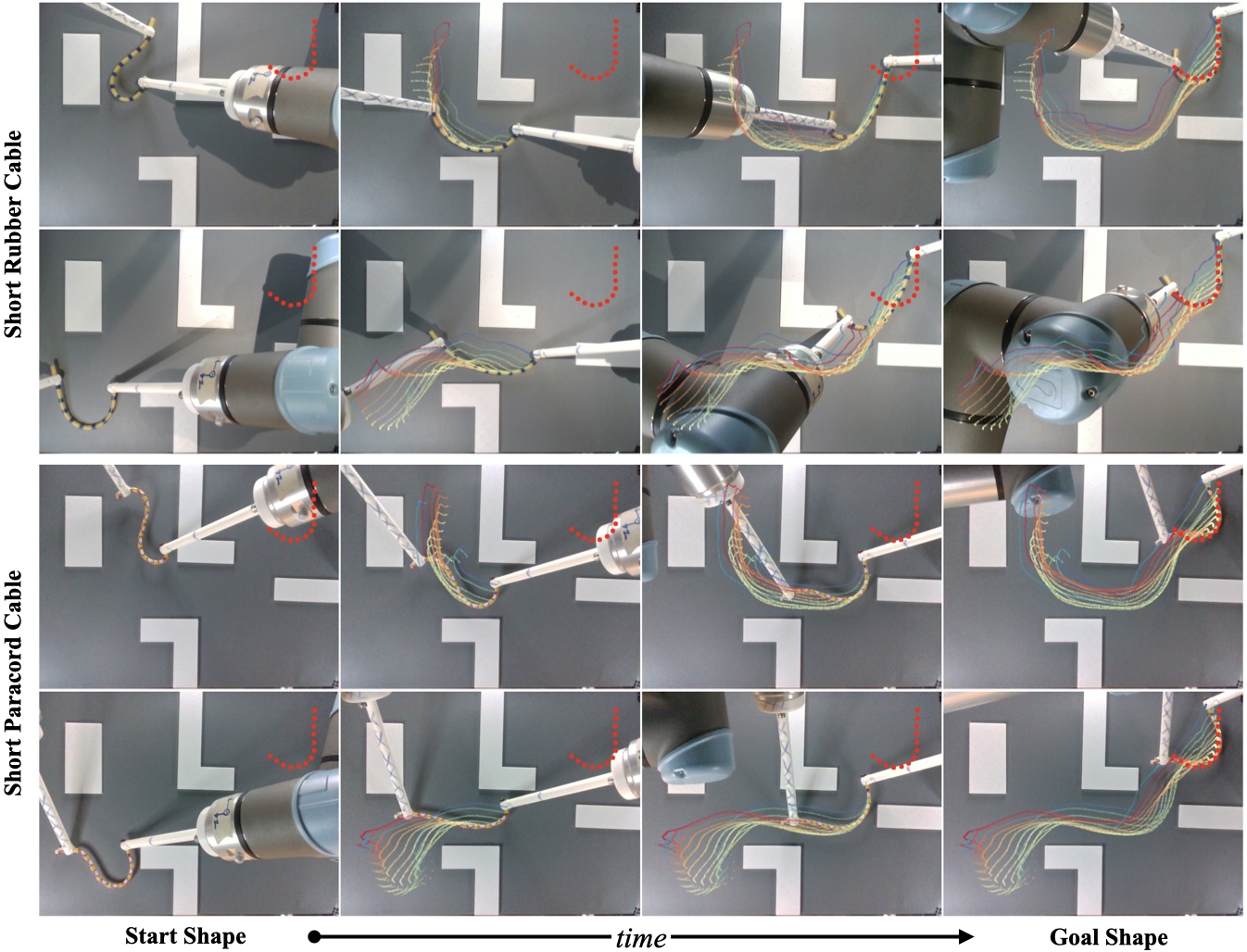}}%
    \caption{(a) Snapshots of multiple obstacle avoidance on the tested long cables. (b) Snapshots of maze navigation on the short rubber and Paracord cables.}
    \label{fig:real_mo_and_ma_snapshots}
\end{figure*}
\subsubsection{Multiple Obstacle Avoidance} As shown in Fig.~\ref{fig:real_mo_and_ma_snapshots}(a), the cables are manipulated through a narrow obstacle gap to the target. A detailed analysis of the deformation planning is presented as a case study. As shown in Fig. \ref{fig:opt_res}(a), the proposed planner generates a smooth deformation flow from the upper-right start to the lower-left target. The optimized decision variable $\boldsymbol{z}$ is plotted in Fig. \ref{fig:opt_res}(b). Examining the distribution of the optimal path parameters in Fig.~\ref{fig:opt_res}(b), the deformation sequence is divided into three stages: \textit{S1}, \textit{S2}, and \textit{S3}. In \textit{S1}, the path parameter $\boldsymbol{\sigma}_1$ (red) increases slightly, while $\boldsymbol{\sigma}_{13}$ (blue) rises significantly to over $0.5$. The DLO shapes at the start (Fig.~\ref{fig:opt_res}(a)) show that the blue segments advance rapidly into the gap, with red segments following more slowly. This optimized deformation sequence allows the cable to undergo significant deformation in \textit{S1}, passing through the gap part by part while avoiding excessive bending. In contrast, a uniform motion approach would push the entire cable through simultaneously, increasing the risk of over-compression. In \textit{S2}, the differences in $\boldsymbol{\sigma}_i$ increments are small, indicating that all keypoints progressed at a similar rate. This suggests the cable mainly translated through the gap with minimal deformation, as seen in the middle trajectory of Fig.~\ref{fig:opt_res}(a). In \textit{S3}, the blue part reached the target earlier with path factors approaching $1$, and the red part arrived slightly afterward.
The optimized trajectory guides the cable through the narrow gap with low potential energy.
We also showcase the deformation tracking of USB cable manipulation in Fig. \ref{fig:opt_res}(c). The reference shapes are closely tracked, with the target error and obstacle distance plotted in Fig.~\ref{fig:opt_res}(d). 
The target error was less than $2$ cm and no collision happened with positive clearance. The manipulation snapshots are shown in Fig.~\ref{fig:real_mo_and_ma_snapshots}(a).
\subsubsection{Maze Navigation} Figure~\ref{fig:real_mo_and_ma_snapshots}(b) presents a more complex environment with multiple obstacles distributed throughout the workspace. In this task, the DLOs are required to navigate through maze-like environment to reach their target configurations. Owing to workspace constraints, shorter rubber and Paracord cables of $0.17$ m were used for evaluation. As shown in the first two rows of Fig.~\ref{fig:real_mo_and_ma_snapshots}(b), the rubber cable was initialized in two different configurations. In both cases, the robots successfully manipulated the cable to the target shape located within a confined region of the maze. This demonstrates the robustness to varying initial states. Similarly, the proposed framework successfully completed the navigation task using a more flexible Paracord cable, achieving accurate deformation without collisions. 
{The results of constrained experimental trials are summarized in Table~\ref{tab:exp_res}. An experiment is considered successful if the shape error is below 3 cm and no collision occurs. Compared to idealized simulation settings, real-world experiments introduce keypoint tracking error, greater noise, and action delays. Nevertheless, no qualitative degradation in performance was observed, indicating that the controller functions effectively without relying on precise state estimation or action execution.}
\subsection{{Case Study}}
\begin{figure}
    \centering
    \includegraphics[width=0.48\textwidth]{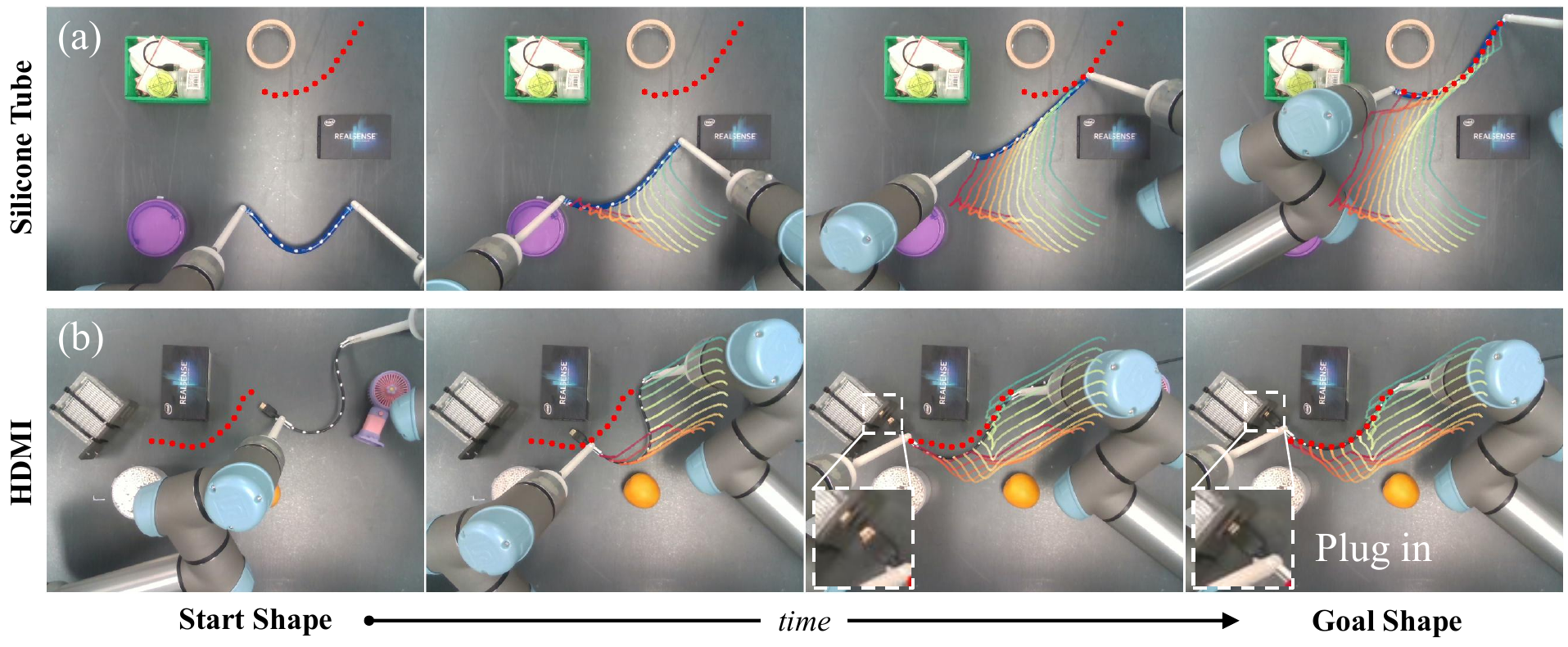}
    \caption{Case study: (a) Zero-shot silicone tube manipulation. (b) Obstacle-aware HDMI cable insertion.}
    \label{fig:case_study}
\end{figure}

{The last part presents application studies that demonstrate the performance and real-world generalization of the proposed framework. In the first case, the objective is to manipulate a soft blue silicone tube to a specified target configuration.}
\begin{table}[tb]
    \centering
    \caption{Real-world Experiment Summary}
    \begin{tabular}{c|ccc}
        \toprule
        Task &Target Error (cm) & Clearance (cm) & Success Rate \\
        \midrule
        Single Obs.   & $1.86 \pm 0.75$ &$1.51 \pm 1.03$ &$16/16$\\
        Multi. Obs.   & $1.58 \pm 0.82$ &$0.77 \pm 0.63$   &$14/16$\\
        Maze Nav.     & $1.94 \pm 0.76$ &$0.43 \pm 0.19$ &$14/16$\\
        \bottomrule
    \end{tabular}
    \label{tab:exp_res}
\end{table}
{This experiment evaluates the zero-shot capability of the neural tracking controller by reusing a deformation model identified from an HDMI cable, without collecting any task-specific data for the tube and relying only on online model adaptation during execution. Despite differences in material properties and geometric dimensions between the HDMI cable and the silicone tube, the proposed approach successfully manipulated the tube to follow the planned deformation sequence and reach the goal, as illustrated in Fig.~\ref{fig:case_study}(a).}
{The second task is to plug an HDMI cable into a NUC computer on a cluttered tabletop. This everyday task requires precise obstacle-aware routing, controlled cable deformation, and accurate endpoint alignment. Figure~\ref{fig:case_study}(b) shows the task setup and execution snapshots, illustrating the robot successfully routing the cable through the clutter and inserting it into the port. The proposed framework effectively plans and tracks obstacle-aware deformations, avoids collisions, and achieves precise endpoint alignment. This case demonstrates the framework’s capability to generate collision-free deformation trajectories while maintaining coherent shape tracking in realistic scenarios.}
\section{Conclusion}
This work presents a novel framework for deformation planning and neural tracking in constrained DLO manipulation. The framework combines a hierarchical deformation planner for global deformation trajectory synthesis, and a neural MPC tracker for obstacle-aware deformation tracking. Furthermore, a Transformer-based deformation model is introduced to enhance MPC tracking by capturing complex robot–DLO interaction dynamics and long-horizon deformation dependencies. Extensive experiments validate the framework's effectiveness across diverse DLOs and constrained manipulation tasks. {Though the proposed approach performs well overall, we found several failures due to two issues. Limited workspace visibility and robot-induced camera occlusions occasionally caused keypoint tracking errors, future work could adopt more robust marker-less DLO state estimation method~\cite{gen_dlo_percep}. The current planner ignores robot configuration constraints, which can cause self-collisions. While reducing dimensionality and remaining robot-agnostic, it may limit performance in complex environments. Future work could integrate joint reasoning over robot kinematics and DLO deformation to improve robustness. Extending the method to real-world scenarios, such as multi-robot cable assembly~\cite{Kejia2025} and human–robot shared manipulation, also presents promising directions.}

\end{document}